\def\year{2022}\relax
\documentclass[letterpaper]{article} 
\usepackage{aaai22}  
\usepackage{times}  
\usepackage{helvet}  
\usepackage{courier}  
\usepackage[hyphens]{url}  
\usepackage{graphicx} 
\usepackage{natbib}  
\usepackage{caption} 
\DeclareCaptionStyle{ruled}{labelfont=normalfont,labelsep=colon,strut=off} 
\usepackage{algorithm}
\usepackage{algorithmic}

\usepackage{newfloat}
\usepackage{listings}
\floatstyle{ruled}
\newfloat{listing}{tb}{lst}{}
\floatname{listing}{Listing}

\usepackage{multirow}
\usepackage{soul}
\usepackage{graphicx}
\usepackage{microtype}
\usepackage{multirow}
\usepackage{amsmath}
\usepackage{amsfonts}
\usepackage{amssymb}
\usepackage{xcolor}
\usepackage[textwidth=0.8in,textsize=tiny]{todonotes}
\usepackage{booktabs}
\usepackage{soul}
\usepackage{tikz}
\usepackage{hhline}
\usepackage{xspace}
\usepackage{subfigure}
\usepackage{url}
\usepackage{enumitem}
\usepackage{cleveref}

\usepackage{array}

\crefformat{section}{\S#2#1#3} 
\crefformat{subsection}{\S#2#1#3}
\crefformat{subsubsection}{\S#2#1#3}

\newcolumntype{H}{>{\setbox0=\hbox\bgroup}c<{\egroup}@{}}
\newcommand{\blue}[1]{{\bf \textcolor{blue}{#1}}}
\newcommand{\gray}[1]{{\it (\textcolor{gray}{#1})}}

\usepackage{mathtools}

\newcommand{\Dmat}{{\bf D}}

\newcommand{\Zmat}{{\bf Z}}

\newcommand{\ie}{\textit{i.e.}}
\newcommand{\eg}{\textit{e.g.}}

\newcommand{\ourmodel}{RetGen\xspace}
\newcommand{\ourtask}{Information-aware Text Generation}{\xspace}

\title{RetGen: A Joint framework for Retrieval and Grounded Text Generation Modeling}

\author{Yizhe Zhang \thanks{Work done at Microsoft Research} \quad\quad Siqi Sun \quad\quad Xiang Gao \quad\quad Yuwei Fang \\ \textbf{Chris Brockett} \quad\quad\textbf{Michel Galley}\quad\quad \textbf{Jianfeng Gao} \quad\quad \textbf{Bill Dolan}\\
    Microsoft Corporation, Redmond, WA, USA  \\
  {\small \tt yizhezhang@fb.com, \{siqi.sun,xiag,yuwfan,mgalley,chrisbkt,jfgao,billdol\}@microsoft.com}
}

\usepackage{bibentry}

\begin{document}

\maketitle

\begin{abstract}
  Recent advances in large-scale pre-training such as GPT-3 allow seemingly high quality text to be generated from a given prompt. However, such generation systems often suffer from problems of hallucinated facts, and are not inherently designed to incorporate useful external information. Grounded generation models appear to offer remedies, but their training typically relies on rarely-available parallel data where information-relevant documents are provided for context. We propose a framework that alleviates this data constraint by jointly training a grounded generator and document retriever on the language model signal. The model learns to reward retrieval of the documents with the highest utility in generation, and attentively combines them using a Mixture-of-Experts (MoE) ensemble to generate follow-on text.  
We demonstrate that both generator and retriever can take advantage of this joint training and work synergistically to produce more informative and relevant text in both prose and dialogue generation.

\end{abstract}

\section{Introduction}
Recent large-scale pre-trained language models (LMs) such as BERT \cite{devlin2018bert}, GPT-2 \cite{Radford2019gpt2}, GPT-3 \cite{Brown2020gpt3}, and T5 \citep{Raffel2019T5} have brought numerous breakthroughs in natural language generation (NLG) across a variety of tasks. These models, however, are not designed to leverage external information to enhance or to verify the predicted text.
\citet{gao2020robust}, for example, demonstrate that they fail to reliably generate responses grounded in real-world knowledge, and may fall short when generating goal-directed responses that are optimized for information-seeking task completion.
These models pose several challenges in information-demanding scenarios: 
First, they are usually trained offline, rendering the model agnostic to the latest information (\eg, asking a chat-bot trained from 2011-2018 about COVID-19).
Second, they are mostly trained on public data,  rendering them less suitable in scenarios where customized or personalized information must be processed (\eg, writing suggestions based on private user-data).  
Third, even in scenarios that call only for public information, generation from these LMs may be unfaithful to the facts (\eg, hallucinations about birth dates), 
especially when the people or entities are less well known and the scenario demands a high degree of fidelity.
As a practical matter, moreover, there remains a fundamental capacity issue in that large LMs cannot effectively represent all the information about every person or entity in the world.
%




A solution that would at first glance seem obvious is to ground the language model in real-world knowledge, which can be present in either structured data (\eg, a knowledge-graph) or unstructured data (\eg, documents such as Wikipedia, user documents or background stories) \cite{Wu2020ControlGround,Ghazvininejad2018AKN,Dinan2018WizardOW,Qin2019CMR}. The advantage of using unstructured grounding data over structured data is that the former provides richer information and it is typically more flexible when it comes to maintaining and updating the information base.
However, training a grounded text generation model that takes additional unstructured documents as input typically demands that the training data contains pairs of context and corresponding oracle documents.
These pairs are seldom available. Recent work, such as REALM~\cite{guu2020realm} and RAG~\cite{lewis2020retrieval}, attempts to leverage information retrieval machinery in real time to mitigate this data paucity in open-domain question answering systems. The approach taken in this paper is in similar vein, but is not confined to the specialized case of question answering, and seeks to present a mechanism to that addresses the broader problem of informational accuracy in text generation. 



In this paper, we investigate the task of generating text by potentially taking advantage from massive reference documents. We called this task as \textit{\ourtask} (ITG). Note that ITG is related but different from open-domain Question Answering (ODQA). In ODQA, the input is typically an information query/question, and the generation is expected to be the answer for that query. In ITG, the input is usually not an information query/question. The task is to potentially leverage any possible external reference documents to predict the next utterance or sentence. Unlike the ODQA, ITG might not always directly seek an answer from retrieved documents, instead it usually requires the retrieved information to be digested as context to subtly influence the generation. Therefore, ITG can be applied to scenarios like dialogue generation and text auto-completion which generalize the open-domain QA scenario.

Below, we present a large-scale general purpose pre-training framework that jointly trains a document retriever and a multi-document grounded generator in end-to-end fashion  and allows these to synergistically cooperate to optimize grounded text generation. 
Our method first selects and scores a collection of documents that are 
most helpful to generation according to the language model signal. The multi-document generator then digests these documents and combines their information according to document-specific attention weights to generate a single prediction in a Mixture-of-Experts (MoE) manner. The main contributions are 
as follows:
\begin{itemize}[wide=0\parindent,noitemsep,topsep=0em]
\item We provide a \textit{joint training framework} for grounded generation and document retrieval with a language model signal.
Our method alleviates the need for oracle parallel data (prose-document pairs) with which to train a grounded model, enabling the use of massive non-parallel corpora.


\item From the \textit{retriever's perspective}, our approach 
uses the language model signal to optimize the retriever, so that the documents with highest utility in generation are returned.

\item From the \textit{generator's perspective}, our approach learns to attend to and combine multiple retrieved documents to achieve a mixture-of-expert-based generation. We apply mutual information maximization (MMI) to further enhance the model.  

\end{itemize}

\vspace{-1mm}
\section{Related Work}
\label{sec:relate}
\paragraph{Retrieval-Augmented Language Modeling}
A series of previous work explores a retrieve-then-edit paradigm for text generation \cite{Peng2019ExemplarDecode, Li2018DelRetGen, Cai2019Match2gen, Hashimoto2018StructPred,yang2019hybrid, song2018ensemble, cai2019skeleton, wu2019response}.
This line of work either directly edits the retrieved text, or feeds the retrieved text to a fixed generator. REALM \cite{guu2020realm} has proposed a Retrieval-augmented encoder to extract salient text span for open-domain QA. The knowledge retriever is pre-trained by leveraging the masked language model signal.
RAG \cite{lewis2020retrieval} fine-tunes models that can leverage the Wikipedia documents to facilitate knowledge-intensive NLP tasks, and achieves strong performance on open-domain QA. Our approach differs in that: 
1) we update the document encoder during training, whereas the document encoder in RAG is fixed. The optimizable document encoder enables us to investigate whether language model signals can be leveraged to improve the document representation learning.
Regularly indexing millions of documents is also technically challenging.
2) we incorporate MMI and retriever correction to further improve the performance.
3) we focus on an information-aware text generation (ITG) task, which is related but different from open-domain QA.
\citet{lewis2020pre} proposed a pre-training objective to reconstruct the original document from retrieved evidence documents, and employ the resulting model to improve translation and summarization results. The bulk of recent work has attempted to perform retrieval-augmented generation to either task-oriented \cite{thulke2021efficient} or open-domain \cite{shuster2021retrieval} response generation. However, their retrievers are not optimized during the training, an thus may be unable to learn from the rich language model signals.



\paragraph{Dense Retrieval Models}
Unlike standard information retrieval techniques such as BM25, Dense Retrieval (DR) models map documents and queries into an embedding space and match them according to semantic similarity. Representative works include \cite{lee2019latent, karpukhin2020dense,luan2020sparse,xiong2020approximate}, which achieve the state-of-the-art performance in tasks like open-domain QA and relevance ranking. Such dense retrieval models can be fine-tuned to accommodate customized needs, and have become a core component in many natural language systems \cite{khandelwal2019generalization, guu2020realm}.

\paragraph{Grounded Generation}
Grounded generation based on external knowledge has been extensively studied. 
Some previous work leverages \textit{structured} external sources like relational knowledge bases \citep{zhu2017flexible, liu2018diffusion} or knowledge graphs \citep{young2018augment} for conversation generation. More recently, \citet{liu2019kgconv} have developed a hybrid graph-path-based method on knowledge graphs augmented with unstructured grounding information. Our work focuses on unstructured (raw text) grounding information and thus avoids the need of pre-constructed knowledge graphs. 
\citet{peng2020soloist}  grounds the task-oriented response generation on the retrieved database states. 

Another line of research exclusively uses the \textit{unstructured} grounding. \citet{Ghazvininejad2018AKN} developed a memory network based model to leverage grounding information from Foursquare. \citet{Dinan2018WizardOW} crowd-sourced conversations where each utterance is grounded in no more than a single sentence. \citet{zhou2018dataset} collected a dataset for grounded conversation generation. \citet{Qin2019CMR} employed a machine reading comprehension (MRC) model to extract salient grounding information to facilitate generation. \citet{Wu2020ControlGround} used a controllable generation framework to generate dialogue responses by applying extracted lexical constraints. 
\citet{zhao2020knowledge} equipped response generation with knowledge selection module. 
Annotated grounding in these works is often ad-hoc and not necessarily optimal for the task. 
Our work differs from these in that we jointly train a retriever and generator to optimize grounded text generation performance, and our proposed model does not rely on annotated text-reference parallel data, with the result that it can be trained on any target dataset without additional annotation.  


\begin{figure*}[ht!]
    \centering
    \includegraphics[scale=0.55]{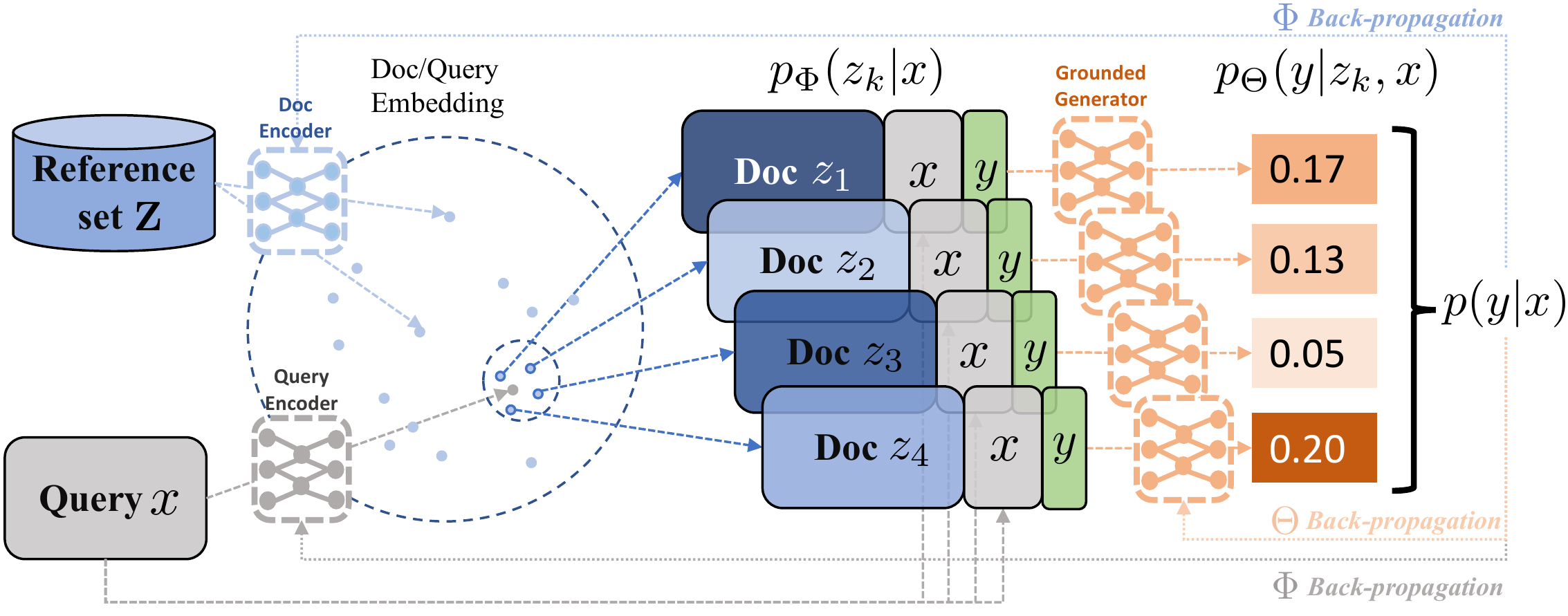}
    \caption{An overview of the \ourmodel framework. A source context query $x$ and documents from a reference database~$\Zmat$ are first mapped to a joint embedding space via different encoders. A Maximum Inner Product Search is performed to retrieve top-$K$ relevant documents ($K=4$ in this figure) with their probability score $p(z_k|x)$. The retrieved documents are separately concatenated with query $x$ and target upcoming text $y$ and passed through a grounded text generator, to compute the document-dependent likelihood $p(y|z_k,x)$. The final objective $p(y|x)$ given by \eqref{eq:final_obj} is optimized to update the retriever parameters $\Phi$ and generator parameters $\Theta$. }
    \vspace{-4mm}
    \label{fig:model}
\end{figure*}

\section{Method}
\subsection{Method Overview}
We begin by formally defining our \textit{\ourtask} (ITG) task and laying out necessary notation. 
ITG aims to predict the upcoming text $y$ that directly follows the existing source prompt $x$ ($x$, $y$ are from a corpus $\Dmat$), while a document reference set $\Zmat$ is accessible. 
In ITG, $\Dmat$ and $\Zmat$ are \textit{non-parallel} to each other. In other words, 
each $x$ is paired with a $y$. However, the association between a document $z$ in $\Zmat$ and the
tuple $(x,y)$ is not necessarily known.


We propose a framework called \textit{\ourmodel} to solve the ITG task. \ourmodel has two components: $i)$ a dense document retriever and $ii)$ a knowledge-grounded text generator. 
The objective of the ITG is to train a model to maximize the likelihood of $y$ given $x$ and $\Zmat$. Formally, it optimizes
\begin{align} \label{eq:obj1}
   p(y|x; \Zmat) = \sum_{z \in \Zmat} p(y|x,z) p(z|x),
\end{align}
In practice, $\Zmat$ often contains millions of documents, rendering enumeration over $z$ impossible. Instead, we leverage a \textit{dense document retriever} $r_\Phi(\cdot)$ to dramatically narrow down the search to a handful relevant documents, where $\Phi$ denotes the retriever parameters. $r_\Phi$ takes $\Zmat$ and $x$ as input and yields relevance scores $\{s_1, \cdots, s_K\}$ of the top-$K$ ($K$ is a hyper-parameter) documents $\tilde{\Zmat} = \{z^{(1)}, \cdots, z^{(K)}\}$.

We further denote the \textit{knowledge-grounded text generator} as $g_\Theta(\cdot)$, where $\Theta$ denotes the generator parameters. This generator module uses $x$ and a single document $z$ as input to produce a probability score for a given reference target $y$, \ie, $g_\Theta(y|x,z) = p(y|x,z)$. The loss can be approximated as:
\begin{align} \label{eq:final_obj}
   \mathcal{L}(\Theta, \Phi) = - \log \sum_{k=1}^K p_\Theta(y|x,z^{(k)}) p_\Phi(z^{(k)}|x),
\end{align}
where $p(z^{(k)}|x) = \exp(s_k)/\sum_{i=1}^K \exp(s_i)$ is the normalized probability (with relevance scores $s$ as logits), and $\tilde{\Zmat} = \{z^{(1)}, \cdots, z^{(K)}\}$ are retrieved from $r_\Phi(\Zmat, x)$. An overview of the model is presented in Figure~\ref{fig:model}.

\subsection{Document Retriever}
For the dense document retriever corresponding to $p_\Phi(\cdot)$ in \eqref{eq:final_obj}, we leverage a model similar to that of \cite{karpukhin2020dense, xiong2020approximate} to achieve efficient document retrieval with sub-linear time. The documents $Z$ and context queries $x$ are mapped into the same dense embedding space. The relevance score $s(x,z)$ is computed as the vector inner product between document embedding $h_z=f_z(z)$ and query embedding $h_x=f_x(x)$, \ie, $s(x,z)=h_x^T h_z$, where $f_z(\cdot)$ and $f_x(\cdot)$ represent learnable encoding networks for document and query respectively. $p(z^{(k)}|x)$ in \eqref{eq:final_obj} is finally given by $\textit{softmax}^{(k)}(s(x,\tilde{\Zmat}))$.

To achieve sub-linear searching time, Maximum Inner Product Search (MIPS) \cite{shrivastava2014asymmetric} is employed. The document embedding vectors are pre-computed and indexed using Locality Sensitivity Hashing (LSH)~\cite{datar2004locality}, so that the query vector can be hashed to a cluster of relatively relevant documents. This search strategy is approximate. However it yields good empirical search results when the document set is large. In practice, we use ANCE \cite{xiong2020approximate} to initialize the retriever. 
 
\subsection{Knowledge-Grounded Text Generator}
\label{sec:generator}
For the knowledge-grounded text generator (GTG) corresponding to $p_\Theta(\cdot)$ in \eqref{eq:final_obj}, we employ a transformer-based architecture akin to GPT-2 \cite{Radford2019gpt2}. The GTG takes one document $z$ and one context query $x$ as the input, and the following text $y$ as the target reference. Specifically, the $z$ and $x$ are first concatenated by a special separator token. The training objective 
follows a standard language model (LM) loss \cite{Radford2019gpt2,zhang2019dialogpt}:
\begin{align} \label{eq:g}
    p_{\Theta}(y|x,z) = \prod_{t=0}^{|y|} p(y_t|x,z,y_{0:t-1}), 
\end{align}
where $y_t$ is the $t$-th token in $y$. $y_{i:j}$ denotes $\{y_i,\cdots,y_j\}$ and $|\cdot|$ denotes the cardinality.  
As opposed to GPT-2, 
we assign different token type embeddings to the tokens in $z$ and $x$ to help the model identify document and context. 

We also employ a distinctive design for the position embedding. The document position id starts with $M$ ($M=400$ in our experiment \footnote{we only select instances with context/response tokens $\leq 256/128$. Thus the maximum number of tokens is around 400.}) while the context position id starts with $0$. The intent is to maximally separate $z$ and $x$, thus reducing the chance that the model will be exposed to hints that $z$ is part of the preceding context.\footnote{GTGs are initialized from GPT-2/DialoGPT, which were trained to recognize tokens with continuous position id as coherent text.} We found this facilitates the model in differentiating the document and the context, and in applying different strategies specific to each.

\subsection{Joint Training}
During the training time, $\Theta$ can be directly optimized from \eqref{eq:final_obj}.
We optimize the objective in \eqref{eq:final_obj} with respect to $\Phi$ by an estimation resembles the Actor-Critic (AC),
\begin{align} \label{eq:rl_estimation}
 \nabla_\Phi  p(y|x) &= \sum_z p(y|z,x) \nabla_\Phi p(z|x) \nonumber\\ 
  &= \sum_z [p(y|z,x)-C] p(z|x) \nabla_\Phi \log p(z|x) 
\end{align}

where the $C$ is a constant baseline. The last step is because $\sum_z \nabla_\Phi p(z|x)\log p(z|x) =  \nabla_\Phi \sum_z p(z|x)  = 0$. $C$ is commonly referred as a ``control variate'' \cite{williams1992simple,nelson1990control} and used to reduce the variance in Monte Carlo estimation as in \eqref{eq:rl_estimation}. The $p(y|z,x)$ can be viewed as the ``return'' in the Actor-Critic algorithm. Document $z$ will receive a positive update if it yields $p(y|z,x)$ larger than to the average performance of the retrieved documents. 
In our experiment, we set $C$ as the expected reward, \ie,  $C=\sum_{z\in \tilde{\Zmat}} p(y|z,x) p(z|x)$. 
Finetuning the retriever model based on \eqref{eq:rl_estimation} needs good initializations from pretrained models to avoid cold-starting. 


Another practical challenge is that all the document embedding vectors need to be \textit{refreshed} once the retriever is updated, which is expensive when the number of documents is large.
Instead of encoding all the documents each time the $\Phi$ is updated to retrieve the top-K document set $\tilde{\Zmat}$, we asynchronously update the retrieved document for every $M$ steps ($M=200$ in our experiments). However, note that even if the $\tilde{\Zmat}$ is fixed for each $K$ steps, the $r_\Phi$ and scores $\{s_1,\cdots,s_K\}$ are still updated at every step.


\subsection{Multi-document Decoding}
\label{sec:decoding}
\paragraph{Mixture-of-Expert (MoE) Decoder}
During the inference time, the retriever first obtains the top-$K$ documents as $\tilde{\Zmat}$, and their corresponding probabilities $p(z|x)$. The generator leverages all document in $\tilde{\Zmat}$ to generator a consensus prediction $\hat{y}$.
One naive approach is to concatenate multiple documents into a ``joint'' document as the input for the generator. The problems for such an approach are that 1) the ``joint'' document may be too long to be efficiently processed;
2) the order of the documents has impact on the generation; 3) the relevance information $p(z|x)$  will be ignored. 

We therefore took a Mixture-of-Expert (MoE) approach following \citet{cho2019MultiDocGen} to decode the model in a document-specific fashion, and ensemble the output distributions at each time step. Specifically, we leverage $K$ copies of the ground text generator $g_\Theta(\cdot)$ trained from \eqref{eq:final_obj}. At time step $t$ 
, we feed each copy of the generator with separate document $z$, the same context $x$, and the same current consensus generation $\hat{y}_{0:t-1}$. We then harvest the individual output distribution from all generator copies, as $\{p(\hat{y}_t^{(1)}),\cdots,p(\hat{y}_t^{(K)})\}$.
The assembled output distribution at step $t$ is finally given by
\begin{align} \label{eq:multi_doc}
 p(\hat{y}_t|x,\hat{y}_{0:t-1}) = \sum_{k=1}^K  p(\hat{y}_t^{(k)}|z^{(k)},x,\hat{y}_{0:t-1}) p(z^{(k)}|x). 
\end{align}

Unlike recent FiD work \cite{izacard2020leveraging}, which ``fuses'' the encoded representations from different documents, our ``fusion'' of information from different document occurs at output token distribution level. FiD requires training a grounded generation model by taking a fixed number of documents as input. However, our MoE approach can directly leverage a grounded generation model trained on single document as input, without additional training or fine-tuning. This yields convenience and flexibility in the number of documents $K$ to leverage for inference. 

We also employ a novel \textit{retriever correction} on the decoder to accommodate the fact that the model is trained to \textit{autoregressively} generate $y$, which implies that the retriever score needs to be updated along the generation. Details are provided in Appendix~\ref{app:ret_cor}.

\paragraph{MMI}
 We further implement a Maximum Mutual Information (MMI) scoring function~\cite{li2015diversity, zhang2018generating} to enhance the ``groundness'' of the generation. MMI employs a pre-trained \textit{backward} grounded text generation model to predict $x$ and $z$ from given prediction $y$, \textit{i.e.}, $p(x,z|y)$.\footnote{This objective is designed to encourage $y$ to incorporate information from both $x$ and $z$.} We first generate a set of hypotheses using top-K sampling. Then we use the probability of $p(x,z|y)$. For multiple $z$ we use the mean of these probabilities to rerank all hypotheses. Intuitively, maximizing backward model likelihood penalizes the bland hypotheses \cite{zhang2019dialogpt} and encourages the generation $y$ to tie better to the input document $z$ and context $x$.



\section{Experimental Setups}
\label{sec:exp}
\paragraph{Datasets}
We use two datasets $\Dmat$, Reddit and arXiv, which cover two information-demanding scenarios (response generation and prose generation)
, to evaluate 
our methods. 


The \textbf{\textit{Reddit}} dataset contains 2.56M/2K/2K training/validation/test conversation instances. The training set is created using the extraction pipeline from the DSTC-7 grounded response generation challenge \cite{galley2019grounded}, which extracts the Reddit threads with time span 2011-2017. 
It contains threads of discussions like a tree, where a reply is a child node to the previous message. Any path along the tree introduces a dialogue session.
Although our approach does not require parallel oracle document, we only select instances that oracle document \footnote{containing URLs linking to Wikipedia domain, see Appendix~\ref{app:data}} can be found.
the reasons are two-fold: $1)$ we hope 
to be able 
to build strong baselines grounded on oracle dataset, which characterize the upper bound performance of a grounded generation model;
$2)$ the oracle documents enable separately evaluating the retriever performance using IR metrics (Recall@$K$). 
We further select test examples from Reddit with time span 2018-2019 by requiring the context to have at least 6 different responses. This yields a 5-reference test set with 2,000 samples. For each instance, one of the 6 human responses is set aside to assess human performance. The average length of the context and response is 44.32 and 14.86 words, respectively.




The \textbf{\textit{arXiv}} dataset is based on \citet{clement2019arxiv}
, which collects a corpus of arXiv articles from 1991 to 2019. We construct the context and target pairs using the abstracts. The final resulting train/validation/test contains 9.6M/57K/2K instances from 1.67M unique abstracts. No parallel oracle document is available for this dataset.



For the \textit{\textbf{reference}} dataset $\Zmat$, we extract about 5.7 million documents from Wikipedia dump of December 2018. For each entry, we only extract the first two paragraphs as these are typically most relevant and summarize the entire document. In addition, we truncate overlong sentences to 100 words, and remove the entry if it contains only one sentence.

More dataset details are provided in Appendix~\ref{app:data}.

\paragraph{Evaluation Metrics}
We performed automatic evaluation using standard machine translation metrics, including BLEU \cite{papineni2002bleu}, METEOR \cite{lavie2007meteor}, and NIST \cite{doddington2002nist}. 
NIST is a variant of BLEU that weights n-gram matches by their information gain, i.e., it indirectly penalizes uninformative n-grams. 
Following ~\citet{zhang2019dialogpt}, we also use Entropy \cite{zhang2018generating} and Dist-n \cite{li2015diversity} to evaluate lexical diversity.
For the Reddit dataset, where 5 references are available for each instance, we compute all relevance metrics and aggregate all of them using max-pooling.

To evaluate how well the predicted text $\hat{y}$ reflects the external information, we propose an evaluation score which we call a \textit{Keyword Matching Ratio (KMR)}.
KMR is defined as
\begin{align} \label{eq:hit}
  &\text{K-words} =  \text{set}(z) \setminus \text{set}(x), \quad  \nonumber \\
  &\text{KMR} = |\text{set}(\hat{y}) \cap \text{K-words}|/|\text{K-words}|,  \nonumber 
\end{align}
where $\cap,\setminus,|\cdot|$ denotes the set intersection, difference and cardinality, respectively. 
For each bag-of-word sets (\ie, $\text{set}(\hat{y}), \text{set}(z) ,\text{set}(x)$), stop words 
based on the python \textsc{nltk} module and frequency in the corpus are removed. 
Intuitively, K-words reflect important information (a set of keywords) in the reference documents $z$ but not covered by context $x$. 
KMR calculates the percentage of these keywords covered by the predicted text $y$. Such a metric assesses the utilization of external information but not the relevance. If a model generates reasonable follow-up text, but fails to incorporate important external information in $z$, KMR will still be low.

\paragraph{Baselines \& Model setups} 
We compared \ourmodel with several baselines in both datasets.
The \textbf{DialoGPT}(345M) \cite{zhang2019dialogpt} and \textbf{GPT-2}(345M) baselines are obtained by finetuning the original pre-trained models on the target training dataset to alleviate the dataset-shifting bias. 
For the Reddit dataset, since we have the oracle document for each conversation, it is possible to train a ground generation model as described in section~\cref{sec:generator} by directly grounding on the oracle documents. This model is denoted as \textit{gDPT} (grounded DialoGPT).
\textbf{gDPT (w/ oracle doc)} and \textbf{gDPT  (w/ random doc)} denote the generation from gDPT model (described in \cref{sec:exp}) using oracle and  random document, respectively. These two models set up the upper and lower bounds of the performance of the grounded generator $g_\Theta$. 

For each dataset, we evaluate 4 variants of \ourmodel: $i)$ \textbf{\ourmodel(K=1)} uses only top-1 retrieved document to generate text; $ii)$ \textbf{\ourmodel(K=4)} uses all top-4 retrieved documents for generation; $iii)$ \textbf{Fixed $\Phi$} is an ablation of \ourmodel(K=4) where the retriever parameters $\Phi$ are frozen during the training; $iv)$ \textbf{+MMI} is a variant of \ourmodel(K=4) using MMI, (\cref{sec:decoding})~\cite{li2015diversity, zhang2018generating}. We first generate 16 hypotheses using top-10 sampling, then select the top hypothesis using reverse model probability of $p(z,x|y)$. The reverse model is also a 345M model fine-tuned from DialoGPT/GPT-2 using the same dataset.


Note that we only perform fine-tuning on existing pre-trained LMs and dense retrievers. All the grounded generators use the same transformer architectures and are initialized with DialoGPT/GPT-2 (345M) weights. The dense retrievers are initialized from ANCE \cite{xiong2020approximate}. 
For the retriever training, we 
index the documents for each $200$ iterations. Models are trained on workstations with 8 Nvidia V100 GPUs. During training, we use $K=4$ for \ourmodel. 

More model setup details
are provided in Appendix~\ref{app:setup}.

\begin{table*}[t!]
\scriptsize
\centering
\resizebox{1.8\columnwidth}{!}{%
\begin{tabular}{r H  r H  r | H r  H r | r| H H H r  |r  r | c |c H H}
	\cmidrule[\heavyrulewidth]{1-20}
	  & \multicolumn{4}{c|}{NIST} & \multicolumn{4}{c|}{BLEU} & MET- & \multicolumn{4}{c|}{Entropy} & \multicolumn{2}{c|}{Dist} & \multicolumn{1}{c|} {Avg. Len.} & \multicolumn{1}{c} {KMR}  \\ 
	Method & N-1 & N-2 & N-3 & N-4 & B-1 & B-2 & B-3 & B-4 & EOR & E-1 & E-2 & E-3 & E-4 &  D-1 &D-2 & & & RT & SCORE \\
	\cmidrule[\heavyrulewidth]{1-20} 
	\multicolumn{20}{c}{\textit{\textbf{Reddit}} dataset} \\
	\cmidrule[\heavyrulewidth]{1-20} 
	DialoGPT & 1.46&	1.59&	1.60&	1.60&	34.78\%&	12.41\%&	5.11\%&	2.34\%&	7.23\%&	5.44&	7.29&	8.01&	8.34&	13.2\%&	32.8\%&	12.0&	- &	-0.958&	0.203\\
	gDPT (w/ oracle doc) & 2.16&	2.37&	2.39&	2.39&	33.84\%&	12.58\%&	5.30\%&	2.57\%&	7.41\%&	5.68&	7.77&	8.65&	9.04&	13.0\%&	33.2\%&	15.1&	4.8\%&	-1.014&	0.202\\
	gDPT (w/ random doc) & 1.87&	2.03&	2.05&	2.05&	29.10\%&	10.14\%&	4.12\%&	1.91\%&	7.12\%&	5.52&	7.63&	8.58&	9.03&	9.9\%&	27.2\%&	18.0&	2.8\%&	-1.064&	0.174\\
	\cmidrule[\heavyrulewidth]{1-20} 
	\ourmodel ($K=1$) & 2.19&	2.39&	2.41&	2.41&	33.67\%&	12.29\%&	5.09\%&	2.32\%&	7.43\%&	5.85&	8.05&	8.95&	9.33&	14.1\%&	37.6\%&	15.6&	4.9\%&	-1.041&	0.188\\
	\ourmodel ($K=4$) & 2.20&	2.40&	2.42&	2.42&	34.03\%&	\textbf{12.53\%}&	5.30\%&	\textbf{2.52\%}&	7.47\%&	5.87&	8.07&	8.98&	9.36&	14.5\%&	38.7\%&	15.3&	5.2\%&	-1.029&	0.191\\
     \ourmodel ($K=4$, Fixed $\Phi$)  & 2.17&	2.37&	2.39&	2.39&	32.54\%&	11.72\%&	4.88\%&	2.31\%&	7.63\%&	5.74&	7.91&	8.82&	9.21&	12.9\%&	34.6\%&	16.9&	4.3\% &	-0.999&	0.184 \\
	  \ourmodel ($K=4$, +MMI) & 2.28&	\textbf{2.44}&	2.46&	\textbf{2.46}&	33.23\%&	10.98\%&	4.19\%&	1.70\%&	\textbf{8.04\%}&	6.56&	9.34&	10.17&	\textbf{10.30}&	\textbf{18.6\%}&	\textbf{60.0\%}&	18.5&	\textbf{6.3\%}&	-1.140&	0.153\\
	\cmidrule[\heavyrulewidth]{1-20} 
	Human oracle & -&	2.13&	2.15&	2.15&	34.96\%&	13.39\%&	6.80\%&	4.25\%&	7.34\%&	7.06&	9.53&	9.93&	9.89&	28.2\%&	77.1\%&	12.9&	5.9\%&	-1.182&	0.145\\
	\cmidrule[\heavyrulewidth]{1-20}
	\cmidrule[\heavyrulewidth]{1-20} 
	\multicolumn{20}{c}{\textit{\textbf{arXiv}} dataset} \\
	\cmidrule[\heavyrulewidth]{1-20} 
	GPT-2 & 0.93&	1.04&	1.06&	1.07&	19.59\%&	9.85\%&	5.91\%&	3.81\%&	8.59\%&	5.92&	8.17&	9.08&	9.34&	20.7\%&	51.3\%&	18.6& - &	-0.684&	0.188\\
	\cmidrule[\heavyrulewidth]{1-20} 
	\ourmodel ($K=1$) & 1.62&	1.81&	1.83&	1.84&	24.12\%&	11.75\%&	6.78\%&	4.19\%&	9.04\%&	5.86&	8.21&	9.26&	9.58&	17.5\%&	46.1\%&	23.6& 3.7\% &	-0.640&	0.169\\
	\ourmodel ($K=4$) & 1.63&	\textbf{1.82}&	1.85&	\textbf{1.86}&	24.22\%&	\textbf{11.85\%}&	6.90\%&	\textbf{4.35\%}&\textbf{9.04\%}&	5.86&	8.21&	9.25&	9.57&	17.5\%&	46.0\%&	23.7& 3.8\% &	-0.638&	0.170\\
	\ourmodel ($K=4$, Fixed $\Phi$) & 1.59&	1.78&	1.81&	1.81&	23.93\%&	11.79\%&	6.89\%&	4.32\%&	9.01\%&	5.86&	8.21&	9.24&	9.56&	17.6\%&	46.4\%&	23.4& 3.7\% &	-0.635&	0.172 \\
	\ourmodel ($K=4$, +MMI) & 1.66&	1.81&	1.83&	1.84&	24.67\%&	10.84\%&	5.81\%&	3.32\%&	8.73\%&	6.34&	8.98&	9.91&	\textbf{10.06}&	19.2\%&	\textbf{59.0\%}&	28.2& \textbf{4.0\%} &	-0.661&	0.115\\
	\cmidrule[\heavyrulewidth]{1-20} 
	Human oracle & - & - & - & - & -& - & - & - & - &6.78&	9.30&	9.91&	9.95&	24.7\%&	71.7\%&	24.4& - & - & -\\
	\cmidrule[\heavyrulewidth]{1-20}
	\end{tabular}
}
	
	\vspace{-2mm}
\caption{Automatic evaluation results on the Reddit (upper) and arXiv (lower) datasets. gDPT w/ oracle(random) doc denotes a grounded generation model directly using oracle(random) document. Fixed $\Phi$ denotes  only fine-tuning generator parameters $\Theta$ while freezing the initial retriever parameters $\Phi$ in ANCE. +MMI represents post-ranking with MMI.}
\label{tab:auto}
\vspace{-2mm}
\end{table*}

\section{Results}
\label{sec:res}
\paragraph{Generation Evaluation}
The automatic evaluation results are summarized in Table~\ref{tab:auto} (the standard deviation of the results are provided in the Appendix~\ref{app:std}). 
We observe that freezing the retriever to pretrained ANCE yield suboptimal evaluation metrics by comparing \textbf{Fixed $\Phi$} and \textbf{\ourmodel(K=4)}. This implies that retriever fine-tuning is crucial to adapt the retriever to the generation task. Consistent with the observations in \citet{zhang2019dialogpt}, the \textbf{MMI} re-ranking procedure produces more diverse text and achieves higher NIST and METEOR scores, albeit with a slight drop in BLEU. We presume the inconsistency
is because NIST generally rewards more for informative and low-frequency n-grams. Incorporating additional information from retrieved documents presumably makes the generation to be more informative diverse.
On the Reddit dataset, \textbf{\ourmodel(K=4)} achieves 
comparable performance to \textbf{\ourmodel (w/ oracle doc)}, indicating the retrieved documents are of high quality.

We also compute KMR, which evaluates the utility of the external document $z$ for generating text $y$, as described in \cref{sec:exp}. For the Reddit dataset, the KMR for gDPT and the human oracle\footnote{The human oracle
only provides a reference baseline and may not be comparable with the compared systems. 
} is calculated against oracle document. Otherwise, KMR is calculated against the retrieved documents by performing a max-pooling over document-specific KMR.
As expected, \ourmodel with MMI generally achieves the highest KMR, as it explicitly maximizes the mutual information between the documents and the output. For both datasets, \ourmodel with more documents and with trainable retriever achieves a higher KMR. Note that KMR 
may not necessarily be associated with generation quality. However, except for MMI, a higher KMR indicates the model is more effective in leveraging the external document to optimize the LM objectives. 

Note that for some metrics the systems achieve higher score than human oracle.
As discussed in \citet{zhang2019dialogpt}, this observation does not imply that the machine generation achieves human parity, but is presumably an artifact of the randomness of human responses in the data.

\begin{table*}[ht!]
\scriptsize
\centering
\begin{tabular}{p{0.34in}|p{2.7in} |p{3.45in} }
\cmidrule[\heavyrulewidth]{1-3}
 & Reddit dataset& ArXiv dataset \\
\cmidrule[\heavyrulewidth]{1-3} 
Context & TIL: All arcade games imported into North America from 1989 to 2000 had the following FBI slogan included into their attract mode: \blue{Winners Don't Use Drugs}. &  (Title: It from Knot)  \blue{Knot physics} is the theory of the universe that not only unified all the fundamental interactions but also explores the underlying physics of quantum mechanics.
\\
\cmidrule[\heavyrulewidth]{1-3} 
(X)PT  & I'm pretty sure that's the slogan of the game in question. & The theory of the knot is a new approach to quantum mechanics. \\
\cmidrule[\heavyrulewidth]{1-3} 
\ourmodel & I have a feeling a major part of the reason was \blue{Nixon} was in charge during that period of history. & A knot is a \blue{finite} sequence of \blue{discrete quantum states} that represent the \blue{gravitational field} in a quantum theory. \\
\cmidrule[\heavyrulewidth]{1-3} 
Retrieved document(s) & \blue{Winners Don't Use Drugs} is an anti-drug slogan that was included in arcade games ... \blue{The slogan was part of a long-term effort} by the United States in its war on drugs started by President Richard \blue{Nixon} in 1971 \gray{Winners Don't Use Drugs}& In loop quantum gravity, ... , s-knots represent \blue{the quantum states of the gravitational field} \gray{S-knot}

Knots have been used for ... Modern physics demonstrates that the \blue{discrete} wavelengths depend on quantum energy levels. ... the Jones polynomial and its generalizations, called the \blue{finite} type invariants... \gray{History of knot theory}
\\
\cmidrule[\heavyrulewidth]{1-3}
\end{tabular}
\caption{Generated examples for Reddit (left) and arXiv (right). (X)PT denotes DialoGPT/GPT. The relevant parts are \blue{highlighted}, and the title of the most relevant retrieved Wikipedia entries are shown in \gray{article title}. }\label{tab:generation}
\vspace{-3mm}
\end{table*}

\paragraph{Generated examples}
We provide generated examples for both datasets in Table~\ref{tab:generation}. The \ourmodel examples are from our best system.
In general, \ourmodel demonstrates ability to integrate information from difference sources including context and multiple references, and sometimes generate text that reflects multi-hop cross-reference among all the sources. We empirically observe that the retrieved documents are usually relevant and may cover orthogonal aspects of the topics in the context. We also visualize how the document attention weights $p(z|x,y_{0:t-1})$ change during the generation process in Appendix~\ref{app:att}. We observed that the attention distribution over documents generally becomes more peaked over steps of generation, indicating the model become more certain as generation proceeds.

Nevertheless, we observe several \textit{failure modes} in our experiments with \ourmodel: $i)$ the retrieved passage may not always be relevant and correct. We find that the \ourmodel can learn to be inclined to avoid using irrelevant documents, but we still see cases where poorly retrieved documents result in incorporation of hallucinations or irrelevant information in final generation; $ii)$ the retrieved passage is relevant, but the grounded generation model may miss correct information and incorporate similar but incorrect/irrelevant information in the generated text (\eg, when asked about who Barack Obama's grandfather was, the system offers his father's name which is also in the retrieved document). These issues do not dominate in our experiments, but resolving them is important and warrants further investigation. 
We provide examples of above problems in Appendix~\ref{app:issue}.

\paragraph{Impact of number of documents $K$} 
From Table~\ref{tab:auto}
, \ourmodel with $K=4$ consistently obtains higher NIST, BLEU and METEOR scores compared with $K=1$, indicating that incorporating multiple retrieved documents may provide better coverage of the references. We also evaluate \ourmodel with $K$ ranging from $1$ to $4$ (Appendix~\ref{app:k}), demonstrating monotonic improvement 
when $K$ increases. We observe no significant improvement with $K>4$.

\begin{figure}[ht!]
    \centering
    \hspace{-3mm}
    \subfigure[Recall@10 (Reddit)]{\label{fig:rl_recall_10}
    \includegraphics[width=0.22\textwidth]{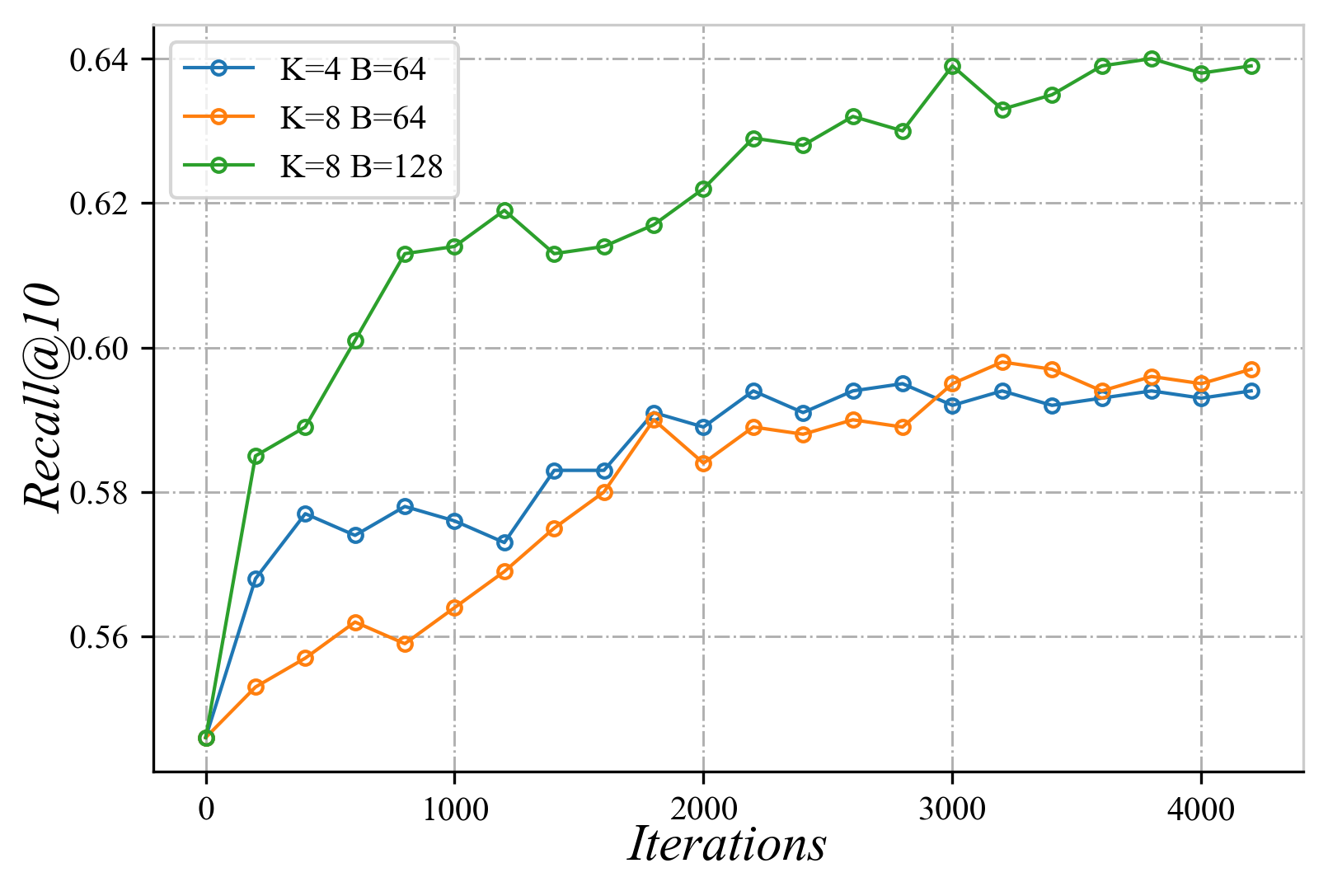}}
    \hspace{-3mm}
    \subfigure[Reward (arXiv)]{\label{fig:rl_reward}
    \includegraphics[width=0.17\textwidth]{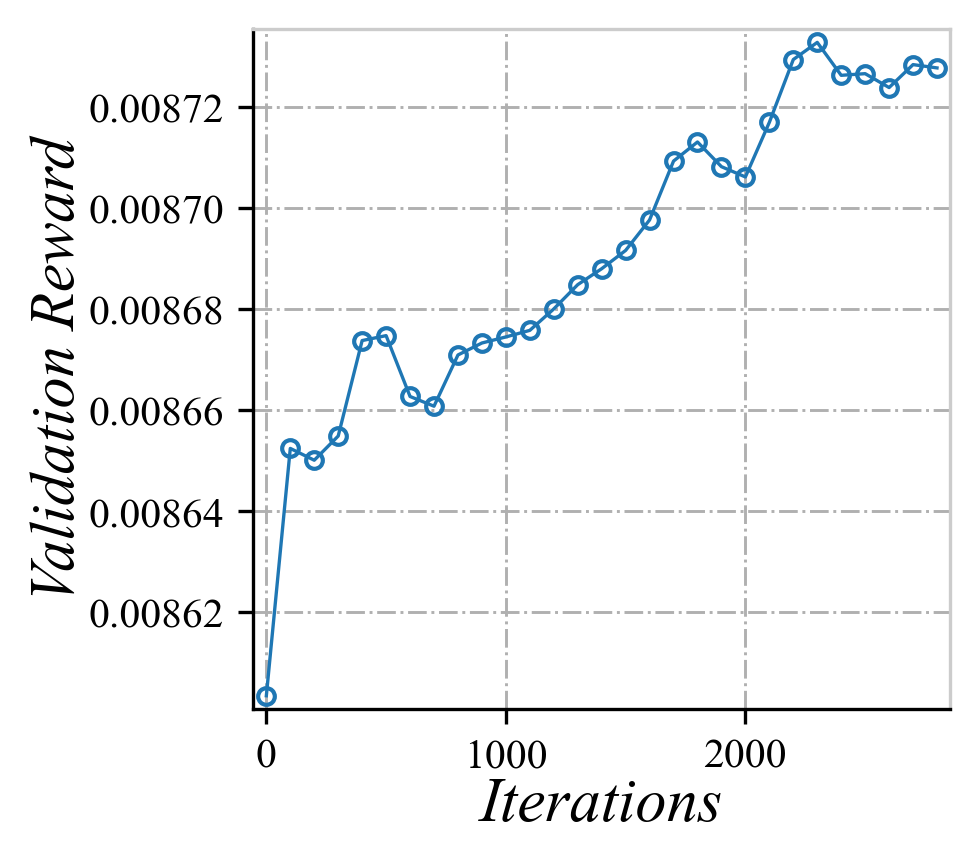}}
    \vspace{-3mm}
    \caption{Recall/Reward on validation set can improve during retriever-only training.
    }
    \vspace{-2mm}
    \label{fig:rl}
\end{figure}

\begin{table}[ht!]
\scriptsize
\centering
\resizebox{1.02\columnwidth}{!}{
\begin{tabular}{r r | c| r l || r r | c| r l}
\cmidrule[\heavyrulewidth]{1-10}
\multicolumn{5}{c||}{Reddit} & \multicolumn{5}{c}{arXiv} \\ 
\cmidrule[\heavyrulewidth]{1-10}
\multicolumn{2}{c|}{System A} & Neutral & \multicolumn{2}{c||}{System B} & \multicolumn{2}{c|}{System A} & Neutral & \multicolumn{2}{c}{System B}\\ 
\cmidrule[\heavyrulewidth]{1-10} 
 \multicolumn{10}{c}{\textbf{Coherence}: \textit{A and B, which is more relevant to, and coherent with the context?}}\\
\cmidrule{1-10}
\ourmodel & \textbf{43.7}\%& 28.3\%& 28.0\%  &   DialoGPT *  & \ourmodel & \textbf{32.1}\%& 41.7\%& 26.3\%  & GPT-2\\
\ourmodel & 33.3\%& 28.6\%& \textbf{38.1}\%  &   MMI & \ourmodel & 29.9\%& 38.7\%& \textbf{31.5}\%  & MMI  \\
\cmidrule{1-10}
\ourmodel & \textbf{40.9}\%& 22.9\%& 36.3\%  &   Human * & \ourmodel & \textbf{34.9}\%& 35.2\%& 29.9\%  & Human * \\
MMI & \textbf{45.9}\%& 23.1\%& 31.0\% &   Human * & MMI  & \textbf{34.9}\%& 35.8\%& 29.3\%  & Human \\

\cmidrule[\heavyrulewidth]{1-10}
 \multicolumn{10}{c}{\textbf{Informativeness}: \textit{A and B, which is more informative (usually more specific content)?}}\\
 \cmidrule{1-10}
\ourmodel & \textbf{44.5}\%& 27.8\%& 27.7\%  &   DialoGPT & \ourmodel & \textbf{36.3}\%& 37.2\%& 26.5\%  & GPT-2 \\
\ourmodel & 32.7\%& 28.3\%& \textbf{39.0}\%  &   MMI  & \ourmodel & 28.9\%& 37.9\%& \textbf{33.2}\%  & MMI  \\
\cmidrule{1-10}
\ourmodel & \textbf{41.1}\%& 21.5\%& 37.5\%  &   Human & \ourmodel & 33.2\%& 32.4\%& \textbf{34.4}\%  & Human  \\
MMI & \textbf{47.5}\%& 21.1\%& 31.4\% &   Human * & MMI & \textbf{34.2}\%& 34.7\%& 31.1\%  & Human \\
\cmidrule[\heavyrulewidth]{1-10}
 \multicolumn{10}{c}{\textbf{Human-likeness}: \textit{A and B, which is more likely to be generated by human rather than a machine?}}\\
 \cmidrule{1-10}
\ourmodel & \textbf{36.4}\%& 34.0\%& 29.6\%  &   DialoGPT  & \ourmodel & \textbf{29.7}\%& 43.6\%& 26.7\%  & GPT-2\\
\ourmodel & 31.3\%& 33.9\%& \textbf{34.9}\%  &   MMI  & \ourmodel & \textbf{28.6}\%& 42.9\%& 28.5\%  & MMI \\
\cmidrule{1-10}
\ourmodel & \textbf{40.1}\%& 28.5\%& 31.4\%  &   Human *  & \ourmodel & \textbf{33.7}\%& 38.9\%& 27.5\%  & Human * \\
MMI & \textbf{40.5}\%& 28.3\%& 31.1\%  &   Human * & MMI & \textbf{33.1}\%& 38.3\%& 28.7\%  & Human \\
\cmidrule[\heavyrulewidth]{1-10} 
\vspace{-3mm}
\end{tabular}
}
\caption{Results of {\bf Human Evaluation} for coherence, informativeness and human-text possibility, showing preferences (\%) for our model (\ourmodel) vis-a-vis baselines and real human responses. \textbf{\ourmodel} denotes \ourmodel with $K=4$, and \textbf{MMI} represents \ourmodel with MMI.
Numbers in bold indicate the preferred systems. 
Statistically significant results with p-value $\leq$ 0.05 are indicated by *. 
}\label{tab:human_eval}
\vspace{-3mm}
\end{table}

\paragraph{Retrieval Evaluation}


From Table~\ref{tab:auto}, it can be seen that optimizing retriever leads to better automatic metrics compared with  generator-only training, indicating that the retriever can benefit from language model signal. 
However, evaluating the retriever improvement using generation metrics as in Table~\ref{tab:auto} is implicit, as the retriever evaluation and generation evaluation are coupled together. To explicitly assess how the retriever can benefit from joint training, we freeze the generator parameters $\Theta$ and only finetune the retriever parameters $\Phi$, and monitor the training process of retriever using either ranking metrics or expected reward in \eqref{eq:rl_estimation}. 

For the Reddit dataset, since oracle documents are available, we monitor the progress of recall@10 during this retriever-only training. The recall value is computed by averaging over 2,000 validation examples. The total number of passage candidates is 10k, including 2k oracle documents for each instance, and 8k hard-negative documents that are close to these 2k oracle documents according to BM25. The results are provided in Figure~\ref{fig:rl} (a). With fluctuation, the recall generally improves as training progresses. Increasing the number of documents $K$ from $4$ to $8$ brought only marginal gain in recall. However, increasing the number of examples in each batch led to more significant improvement of the recall.  

For the arXiv dataset, since recall cannot be computed, we instead monitor the expected reward/return ($r=\sum_z p(y|z,x) p(z|x)$) over 2,000 validation examples. Our reasoning here is that with fixed $\Theta$, if the reward can be improved (\ie, the target $y$ is more likely given the current $z$), the only possible explanation is that the retrieved documents are more relevant and helpful in predicting the oracle target. 
We observed that this reward metric can to some extent improve as training progresses (Figure~\ref{fig:rl} (b)). This verifies that the retriever is being optimized and benefits from LM signals.


\paragraph{Human Evaluation}
Overall judge preferences in each of the 3 categories are shown in Table \ref{tab:human_eval}, where the 5-point scale has been collapsed to a 3-point preference for clarity. A moderate preference can be observed for the variant of \ourmodel with MMI over vanilla \ourmodel.
Table \ref{tab:human_eval} suggests that the \ourmodel may begin to approximate human quality. As has been observed elsewhere, e.g., \citet{zhang2019dialogpt}, we found that judges often prefer model generation over human responses. In the case of the Reddit dataset, we speculate that the original human responses may be more erratic and idiosyncratic than system outputs. Human evaluation of the arXiv dataset, meanwhile, is intrinsically difficult as responses typically involve domain knowledge: human judges may prefer system generated text that is potentially easier to understand.\footnote{This is consistent with the findings in \citet{freitag2021experts} for MT to the effect that crowd-sourced human evaluation is error-prone and may not be as accurate as some automatic metrics.} How to evaluate generated text as systems improve remains a challenge, but further exploration of these issues falls beyond the scope of this work.  
%
Further details, including the human evaluation template used, are provided in the Appendix~\ref{app:human}. 


\section{Conclusion}

We present a joint training framework to simultaneously optimize a dense passage retriever and a knowledge-grounded text generator in an end-to-end fashion. This approach enables leveraging the LM signal to optimize the information retrieval sub-component and thus permits the generation pipeline to output more informative text.  
The resulting algorithm leverages multiple retrieved documents during decoding time and generates text by selectively summarizing and combining information collected from all the references. We have demonstrated the effectiveness of this algorithm via crowd-sourced human evaluation and automatic evaluation that uses generation and retrieval metrics. 
In future work, 
we plan also to leverage QA and cloze task objectives for factuality evaluation \cite{eyal-etal-2019-question, huang-etal-2020-knowledge}. 
We discuss the ethical impact of this work in Appendix~\ref{app:bi}.


\bibliography{rag}

\appendix

\onecolumn

\begin{center}
    {\Large \bf Appendix for RetGen: A Joint framework for Retrieval andGrounded Text Generation Modeling}
\end{center}

\section{Broader Impact}
\label{app:bi}
This work focuses on improving the natural language processing (NLP) and general artificial intelligence (AI) research community. Our work can be leveraged to improve natural language generation (NLG) models, including but not limited to text editing, conversational agents and question answering systems. 
The \textbf{broader impact} the \textbf{risks} of this work are summarized as following:
\begin{itemize}[wide=0\parindent,noitemsep,topsep=0em]
    \item 
    This work can facilitate research in the NLG tasks in a generic manner, to potentially reduce hallucinations and offensive generations in applications like virtual assistants. 
    \item this work is a fundamental research work that focuses on the technical improvement, thus we have NOT imposed additional aggressive filtering techniques to the text data we used, beyond what has been performed to the original dataset from their sources. The text data we used may have offensiveness/toxicity/fairness/bias issues that we haven't been able to identify, as those are not the main focus of this work. 
    \item Given above potential risk, due to the nature of natural language generative models, we note that the generations or outputs of this work, though not likely, may reflect gender and other historical biases implicit in the data. Under rare circumstances, the generations may exhibit a mild extent of unethical, biased, or offensive attitude. These are known issues with current state-of-the-art text generation models. We would hope that a better grounded text generation system as what we present can enable further mitigation strategies to inappropriate and hallucinated generations.
\end{itemize}

\section{Retriever Correction}
\label{app:ret_cor}
The fact that the model is trained to \textit{autoregressively} generate $y$ implies that the retriever score needs to be updated along the generation. To see this, we revisit \eqref{eq:final_obj} by expanding the $p(y|x)$ as $p(y_0|x)\prod_{t=1}^{|y|} p(y_t|x, y_{0:t-1})$, where 
\begin{align}
   & p(y_t|x, y_{0:t-1})  = \sum_{z} p(y_t|x,z,y_{0:t-1}) p(z|x,y_{0:t-1}) \nonumber \\
   &= \sum_{z}   p(y_t|x,z,y_{0:t-1})  p(z|x) p(y_{0:t-1}|z,x) / p(y_{0:t-1}|x). \nonumber
\end{align}

Note that the first term $p(y_t|x,z,y_{0:t-1})$ can be directly obtained from the generator model in \eqref{eq:g}. 
The term $p(y_{0:t-1}|z,x) / p(y_{0:t-1}|x)\triangleq F_t$ serves as a \textit{correction factor} for updating the retriever's belief for document relevance with newly seen/generated text $y_{0:t-1}$. It can be computed by $F_t=p(y_{0:t-1}|z,x) / \sum_{z'} p(y_{0:t-1}|z',x)p(z'|x)$. When this factor is greater than 1 (\ie, being grounded on $z$ improves the probability of $y_{0:t-1}$), the corresponding $z$ will be assigned with a higher probability. The computation cost of the $F_t$ is negligible. The retriever correction simply multiplies this correction factor $F_t$ to 
\eqref{eq:multi_doc} at each time step.

\section{Dataset details}
\label{app:data}
We use two datasets $\Dmat$, \textit{\textbf{Reddit}} and \textit{\textbf{arXiv}}, which cover response generation and prose generation respectively, to evaluate the effectiveness of the proposed methods. 

For \textbf{Reddit} dataset, the training data is created using a pipeline similar to that in the DSTC-7 grounded response generation challenge \cite{galley2019grounded}: We first select the Reddit discussion threads that contain urls in the description, crawled from the Reddit with time span 2011-2017.
Then, we restrict the url domain to Wikipedia, and extract the linked oracle passage by selecting the most relevant passage to the context according to ROUGE-F1. This yields about 2.56M data instances.

For \textbf{ArXiv} dataset, 
for each sentence 
in an abstract, we use the preceding sentences as the context for predicting the current sentence. 
we consider the title to be the first sentence.
We processed the texts to replace citations, urls, and equations by special tokens. We apply a filtering step to select instances in the test set that are likely to benefit from external information by using Tagme \citep{tagme}, a Wikipedia name entity recognition (NER) tool. Tagme identifies the named entities that exist as Wikipedia entries and meanwhile occur in the target sentence.
The Tagme threshold, which balances the precision and recall of NER, is set at $0.7$. We only retain instances that pass this threshold.
The final resulting train/validation/test contains 9.6M/57K/2K instances from 1.67M unique abstracts.


\section{Additional details of experimental setup}
\label{app:setup}
For the retriever training, we save model checkpoints and index the documents for each $200$ iterations.
We observe that reducing the indexing frequency to $100$ provides marginal performance gain, while yielding more computation.
We set the learning rate to be $10^{-6}$ and batch size to be $128$ for most of the experiments. 
During training, we use $K=4$ for \ourmodel. We only test the $K$ up to $4$ due to GPU memory constraint. All generations except for MMI use greedy decoding. 
All compared models are trained until no progress can be observed on validation loss. Models are trained on workstations with 8 Nvidia V100 GPUs.

\section{Impact of number of document K}\label{app:k}
Below are the automatic evaluation results using different $K$ during the decoding time in Table~\ref{tab:auto_num_docs}, for both Reddit and arXiv datasets. It can be seen that, in general, incorporating more documents yields better automatic metrics.
\begin{table*}[ht!]
\scriptsize
\centering
\resizebox{1.0\columnwidth}{!}{%
\begin{tabular}{r H  r H  r | H r  H r | r| H H H r  |r  r | c  H H H}
	\cmidrule[\heavyrulewidth]{1-20}
	\bf{Reddit} dataset & \multicolumn{4}{c|}{NIST} & \multicolumn{4}{c|}{BLEU} & MET- & \multicolumn{4}{c|}{Entropy} & \multicolumn{2}{c|}{Dist} & \multicolumn{1}{c} {Avg. Len.} & \\ 
	Method & N-1 & N-2 & N-3 & N-4 & B-1 & B-2 & B-3 & B-4 & EOR & E-1 & E-2 & E-3 & E-4 &  D-1 &D-2 & & & RT & SCORE \\
	\cmidrule[\heavyrulewidth]{1-20} 
	\ourmodel ($K=1$) & 2.19&	2.39&	2.41&	2.41&	33.67\%&	12.29\%&	5.09\%&	2.32\%&	7.43\%&	5.85&	8.05&	8.95&	9.33&	14.1\%&	37.6\%&	15.6&	2.4\%&	-1.041&	0.188\\
	\ourmodel ($K=2$) & 2.19&	2.39&	2.40&	2.40&	33.65\%&	12.25\%&	5.08\%&	2.36\%&	7.49\%&	5.83&	8.04&	8.95&	9.33&	14.0\%&	37.4\%&	15.6&	2.3\%&	-1.042&	0.190\\
	\ourmodel ($K=3$) & 2.20&	2.39&	2.41&	2.41&	34.05\%&	12.53\%&	5.24\%&	2.47\%&	7.46\%&	5.86&	8.06&	8.96&	9.34&	14.5\%&	38.5\%&	15.3&	2.2\%&	-1.037&	0.191\\
	\ourmodel ($K=4$) & 2.21&	2.41&	2.42&	2.43&	32.88\%&	12.04\%&	5.00\%&	2.31\%&	7.67\%&	5.81&	8.02&	8.95&	9.35&	13.2\%&	36.1\%&	16.5&	2.2\%&	-1.024&	0.184\\
	\cmidrule[\heavyrulewidth]{1-20}
	\end{tabular}
}
\resizebox{1.0\columnwidth}{!}{%
\begin{tabular}{r H  r H  r | H r  H r | r| H H H r  |r  r | c H H}
	\cmidrule[\heavyrulewidth]{1-19}
	\bf{ArXiv} dataset & \multicolumn{4}{c|}{NIST} & \multicolumn{4}{c|}{BLEU} & MET- & \multicolumn{4}{c|}{Entropy} & \multicolumn{2}{c|}{Dist} & \multicolumn{1}{c} {Avg. Len.}  \\ 
	Method & N-1 & N-2 & N-3 & N-4 & B-1 & B-2 & B-3 & B-4 & EOR & E-1 & E-2 & E-3 & E-4 &  D-1 &D-2 & & RT & SCORE\\
	\cmidrule[\heavyrulewidth]{1-19} 
	\ourmodel ($K=1$) & 1.62&	1.81&	1.83&	1.84&	24.12\%&	11.75\%&	6.78\%&	4.19\%&	9.04\%&	5.86&	8.21&	9.26&	9.58&	17.5\%&	46.1\%&	23.6&	-0.640&	0.169\\
	\ourmodel ($K=2$) & 1.62&	1.81&	1.84&	1.85&	24.09\%&	11.82\%&	6.90\%&	4.33\%&	9.03\%&	5.86&	8.21&	9.26&	9.57&	17.4\%&	46.0\%&	23.7&	-0.633&	0.171\\
	\ourmodel ($K=3$) & 1.62&	1.82&	1.84&	1.85&	24.25\%&	11.91\%&	6.93\%&	4.35\%&	9.13\%&	5.87&	8.21&	9.25&	9.57&	17.5\%&	46.2\%&	23.5&	-0.634&	0.171\\	
	\ourmodel ($K=4$) & 1.63&	1.82&	1.85&	1.86&	24.22\%&	11.85\%&	6.90\%&	4.35\%&	9.04\%&	5.86&	8.21&	9.25&	9.57&	17.5\%&	46.0\%&	23.7&	-0.638&	0.170\\
	\cmidrule[\heavyrulewidth]{1-19}
	\end{tabular}
}
	\vspace{-2mm}
\caption{Automatic evaluation results on the Reddit (upper) and arXiv (lower) datasets with different numbers of retrieved document for decoding time}\label{tab:auto_num_docs}
\vspace{-2mm}
\end{table*}

\section{Standard deviation of automatic metrics}
\label{app:std}
We estimate the standard deviation based on bootstrapping the test set (80\% resampling ratio without replacement, 10 sets for each method), the results are provided in table~\ref{tab:std}. All the pairwise comparisons are significant (p<0.0001) based on the unpaired two-sided t-test (with Bonferroni correction).

\begin{table*}
    \centering
    \scriptsize
    \resizebox{1.0\columnwidth}{!}{
    \begin{tabular}{|l|l|l|l|l|l|l|l|l|l|l|}
    \hline
         & NIST2 & NIST4 & BLEU2 & BLEU4 & METEOR & entropy4 & dist-1 & dist-2 & Avg. Length & KMR \\ \hline
        Reddit dataset  &  &  &  &  &  &  &  &  &  &  \\ \hline
        DialoGPT & 1.590$\pm$0.003 & 1.602$\pm$0.004 & 12.412$\pm$0.000 & 2.341$\pm$0.000 & 7.235$\pm$0.000 & 8.336$\pm$0.006 & 13.153$\pm$0.001 & 32.803$\pm$0.001 & 12.027$\pm$0.051 & - \\ \hline
        gDPT (w/ oracle doc) & 2.374$\pm$0.004 & 2.393$\pm$0.004 & 12.582$\pm$0.001 & 2.575$\pm$0.000 & 7.411$\pm$0.000 & 9.039$\pm$0.005 & 12.962$\pm$0.001 & 33.238$\pm$0.002 & 15.128$\pm$0.069 & 4.782$\pm$0.181 \\ \hline
        gDPT (w/ random doc) & 2.033$\pm$0.004 & 2.051$\pm$0.004 & 10.143$\pm$0.000 & 1.913$\pm$0.000 & 7.119$\pm$0.000 & 9.031$\pm$0.006 & 9.887$\pm$0.001 & 27.231$\pm$0.002 & 18.016$\pm$0.059 & 2.764$\pm$0.148 \\ \hline
        RetGen($K=1$) & 2.392$\pm$0.004 & 2.406$\pm$0.004 & 12.292$\pm$0.001 & 2.324$\pm$0.000 & 7.432$\pm$0.000 & 9.329$\pm$0.005 & 14.132$\pm$0.001 & 37.583$\pm$0.002 & 15.606$\pm$0.053 & 4.887$\pm$0.159 \\ \hline
        RetGen($K=4$) & 2.404$\pm$0.005 & 2.419$\pm$0.004 & 12.526$\pm$0.000 & 2.525$\pm$0.000 & 7.468$\pm$0.000 & 9.361$\pm$0.007 & 14.475$\pm$0.001 & 38.654$\pm$0.002 & 15.306$\pm$0.063 & 5.175$\pm$0.192 \\ \hline
        Fixed $\Phi$ & 2.367$\pm$0.004 & 2.391$\pm$0.005 & 11.721$\pm$0.001 & 2.306$\pm$0.001 & 7.633$\pm$0.000 & 9.210$\pm$0.005 & 12.873$\pm$0.001 & 34.645$\pm$0.002 & 16.865$\pm$0.069 & 4.297$\pm$0.208 \\ \hline
        +MMI & 2.435$\pm$0.004 & 2.458$\pm$0.004 & 10.984$\pm$0.001 & 1.699$\pm$0.000 & 8.039$\pm$0.000 & 10.302$\pm$0.007 & 18.593$\pm$0.001 & 60.021$\pm$0.002 & 18.491$\pm$0.059 & 6.321$\pm$0.183 \\ \hline
        Human oracle & 2.128$\pm$0.005 & 2.150$\pm$0.005 & 13.386$\pm$0.001 & 4.250$\pm$0.000 & 7.345$\pm$0.000 & 9.888$\pm$0.007 & 28.230$\pm$0.001 & 77.121$\pm$0.002 & 12.891$\pm$0.054 & 5.935$\pm$0.152 \\ \hline
        \hline
        Arxiv dataset &  &  &  &  &  &  &  &  &  &  \\ \hline
        GPT-2 & 1.038$\pm$0.011 & 1.071$\pm$0.010 & 9.855$\pm$0.001 & 3.806$\pm$0.001 & 8.591$\pm$0.000 & 9.337$\pm$0.002 & 20.691$\pm$0.001 & 51.327$\pm$0.001 & 18.643$\pm$0.042 & - \\ \hline
        RetGen($K=1$) & 1.805$\pm$0.010 & 1.837$\pm$0.014 & 11.754$\pm$0.001 & 4.186$\pm$0.000 & 9.038$\pm$0.000 & 9.582$\pm$0.002 & 17.529$\pm$0.001 & 46.081$\pm$0.001 & 23.604$\pm$0.034 & 3.663$\pm$0.119 \\ \hline
        RetGen($K=4$) & 1.822$\pm$0.010 & 1.857$\pm$0.012 & 11.848$\pm$0.001 & 4.355$\pm$0.001 & 9.044$\pm$0.000 & 9.567$\pm$0.002 & 17.479$\pm$0.001 & 46.024$\pm$0.001 & 23.671$\pm$0.031 & 3.787$\pm$0.131 \\ \hline
        Fixed $\Phi$ & 1.780$\pm$0.012 & 1.811$\pm$0.010 & 11.792$\pm$0.001 & 4.318$\pm$0.001 & 9.009$\pm$0.000 & 9.559$\pm$0.002 & 17.641$\pm$0.001 & 46.447$\pm$0.002 & 23.414$\pm$0.037 & 3.738$\pm$0.140 \\ \hline
        +MMI & 1.814$\pm$0.012 & 1.839$\pm$0.011 & 10.840$\pm$0.001 & 3.323$\pm$0.001 & 8.735$\pm$0.000 & 10.061$\pm$0.003 & 19.207$\pm$0.001 & 59.013$\pm$0.001 & 28.210$\pm$0.031 & 3.971$\pm$0.147 \\ \hline
        Human oracle & - & - & - & - & - & 9.952$\pm$0.002 & 24.667$\pm$0.001 & 71.696$\pm$0.002 & 24.384$\pm$0.035 & - \\ \hline
    \end{tabular}
    }
    \caption{Standard deviation of automatic metrics.}\label{tab:std}
\end{table*}

\section{Issues with generation}
\label{app:issue}
Below, we provide examples where the \ourmodel can fail to generate text that are factually correct in either retriever or grounded generator in Table~\ref{tab:generation_fail}. Even if these generation outputs are not very common, they still pose important challenges for improving the models. 

\begin{table*}[ht!]
\centering
\scriptsize
\small
\begin{tabular}{p{0.5in}|p{2.2in} |p{2.2in} }
\cmidrule[\heavyrulewidth]{1-3}
 & Retrieved passage is less relevant & Passage is correct, however the model fails to incorporate the right information\\
\cmidrule[\heavyrulewidth]{1-3} 
Context &  TIL Chess Grandmaster Bobby Fischer died at the Age of 64. There are 64 squares on a chess board. &   TIL the oldest naval ship still in commission is HMS Victory. She was launched in 1765.
\\
\cmidrule[\heavyrulewidth]{1-3} 
\ourmodel & And if you ever play chess, you will realize that a board is \blue{a board of 64 pieces}, with one king, one queen, two knights. It is very difficult to play with a board that has so many pieces & It's not in commission anymore. The Royal Navy has retired her \blue{since 1824 because the crew is dead}. It's now a museum. \\
\cmidrule[\heavyrulewidth]{1-3} 
Retrieved document & The following outline is provided as an overview of and topical guide to chess: Chess is a two-player board game played on a chessboard (a square-checkered board with 64 squares arranged in an eight-by-eight grid). In a chess game, each player begins with sixteen pieces: one king, one queen, two rooks, two knights, two bishops, and eight pawns. The object of the game is to checkmate the opponent's king, whereby the king is under immediate attack (in "check") and there is no way to remove or defend it from attack, or force the opposing player to forfeit \gray{Outline of Chess} & HMS "Victory" is a 104-gun first-rate ship of the line of the Royal Navy, ordered in 1758, laid down in 1759 and launched in 1765. She is best known for her role as Lord Nelson's flagship at the Battle of Trafalgar on 21 October 1805. She additionally served as Keppel's flagship at Ushant, Howe's flagship at Cape Spartel and Jervis's flagship at Cape St Vincent. After \blue{1824}, she was relegated to the role of harbour ship. In 1922, she was moved to a dry dock at Portsmouth, England, and preserved as a museum ship \gray{HMS Victory}
\\
\cmidrule[\heavyrulewidth]{1-3} 
Issue & The Wiki entry for Bobby Fischer is not the top-1. The generated text deviates from the major topic which is Bobby Fischer. The highlighted text contains hallucination  & The highlighted generation either use wrong part of the retrieved document, or hallucinates facts. 
\end{tabular}
\caption{Failure modes for \ourmodel. }\label{tab:generation_fail}
\end{table*}

\section{Attention Visualization of multi-document MoE decoder}
\label{app:att}
\begin{figure}[ht!]
    \centering
    \includegraphics[scale=0.9]{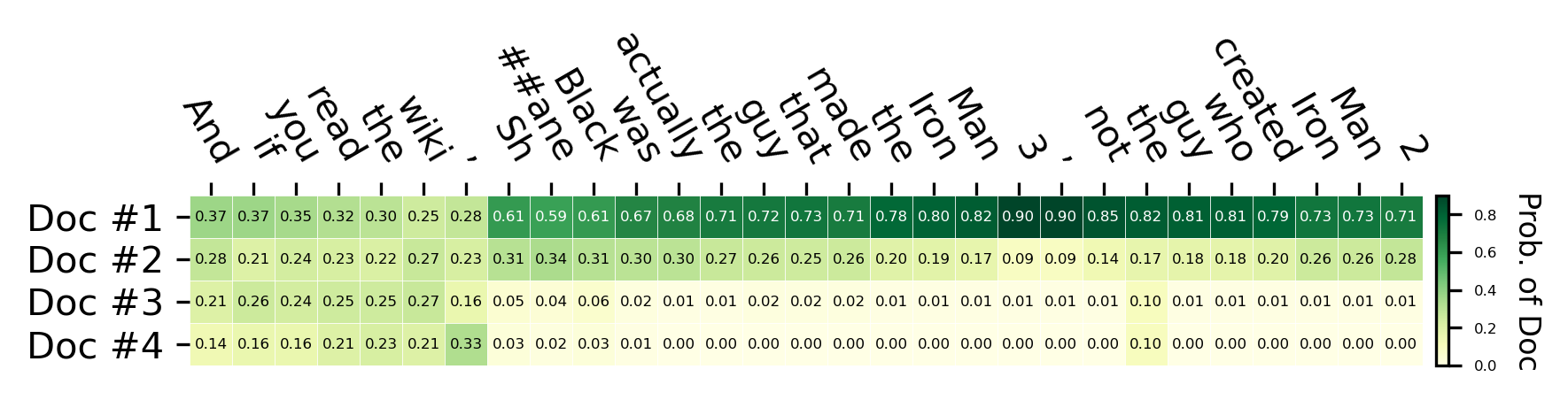}
    \caption{Attention map for multi-document MoE decoder. The context sentence is \textit{TIL the 3 in iron man 3 is a reference to the movies iron man and iron man 2. The usage of the number 3 implies that the movies are in chronological order}, and the resulting MoE generation over 4 documents is \textit{And if you read the wiki, Shane Black was actually the guy that made the Iron Man 3, not the guy who created Iron Man 2}. The retrieved documents are shown in Table~\ref{tab:att_docs}.}
    \label{fig:attention}
\end{figure}

To better understand how the document attention weights $p(z|x,y_{0:t-1})$ are influenced by the documents and existing generated text over the progress of generation, we visualize the attention weights in Figure~\ref{fig:attention}. The context sentence $x$ for the given example is \textit{TIL the 3 in iron man 3 is a reference to the movies iron man and iron man 2. The usage of the number 3 implies that the movies are in chronological order}, and the retrieved top-4 documents are provided in Table~\ref{tab:att_docs}. 

It can be seen that at the initial stage of the generation, the MoE decoder refers to all retrieved documents with relatively even probability. However as the generation become more specific (\eg, mentioning ``Shane Black''), the MoE decoder starts to focus more on the first two documents and assigns negligible attention to the documents \#3 and \#4. We observe that it is typical that during generation, the MoE decoder gradually reinforces the attention to one or two documents by looking at its own existing generation, and the attention distribution becomes more peaked. This typically reduces the likelihood that irrelevant documents (like document \#4 in Table~\ref{tab:att_docs}) will have large impact on generation.

\begin{table*}[ht!]
\centering
\scriptsize
\small
\begin{tabular}{p{1.5in}|p{1.5in} |p{1.5in}|p{1.5in}}
\cmidrule[\heavyrulewidth]{1-4}
Document \#1 & Document \#2 & Document \#3 & Document \#4\\
\cmidrule[\heavyrulewidth]{1-4} 
Studios and distributed by Walt Disney Studios Motion Pictures. It is the sequel to "Iron Man" (2008) and "\blue{Iron Man 2}" (2010), and the seventh film in the Marvel Cinematic Universe (MCU). The film was directed by \blue{Shane Black} from a screenplay he co-wrote with Drew Pearce, and stars Robert Downey Jr. as Tony Stark / Iron Man alongside Gwyneth Paltrow, Don Cheadle, Guy Pearce, Rebecca Hall, Stphanie Szostak, James Badge Dale, Jon Favreau, and Ben Kingsley &  
\blue{Iron Man 2} is a 2010 American superhero film based on the Marvel Comics character Iron Man. Produced by Marvel Studios and distributed by Paramount Pictures, it is the sequel to "Iron Man" (2008) and the third film in the Marvel Cinematic Universe (MCU). Directed by Jon Favreau and written by Justin Theroux, the film stars Robert Downey Jr. as Tony Stark / Iron Man alongside Gwyneth Paltrow, Don Cheadle, Scarlett Johansson, Sam Rockwell, Mickey Rourke, and Samuel L. Jackson &  
\blue{Iron Man 3} (Original Motion Picture Soundtrack) is the film score for the Marvel Studios film, "Iron Man 3" by Brian Tyler, released on April 30, 2013. A separate soundtrack and concept album titled, Iron Man 3: Heroes Fall (Music Inspired by the Motion Picture) by various artists was released on the same date by Hollywood Records and Marvel Music. Composer Brian Tyler acknowledged that the film's score needed to be darker and more melodic than Ramin Djawadi and John Debney's previous scores, citing the change in Tony Stark's life following the events of "The Avengers" as the catalyst &
Men in Black 3 (alternatively Men in Black III, and stylized as "MIB)" is a 2012 American science fiction action comedy film directed by Barry Sonnenfeld and starring Will Smith, Tommy Lee Jones and Josh Brolin. It is the third installment in the "Men in Black" film series which in turn is loosely based on the comic book series "The Men in Black" by Lowell Cunningham. It was released fifteen years after the original "Men in Black" (1997) and ten years after the first sequel, "Men in Black II" (2002)
\\
\end{tabular}
\caption{Retrieved documents for the context \textit{TIL the 3 in iron man 3 is a reference to the movies iron man and iron man 2. The usage of the number 3 implies that the movies are in chronological order}. }\label{tab:att_docs}
\end{table*}

\section{Additional Details of Human Evaluation }
\label{app:human}

Judges were vetted with a simple qualification test and were checked periodically for spamming. Held-out from the human text (for positive instances) and random generated text (for negative instances) were used to provide unambiguous cases for spam detection and training examples. 
Judges were paid \$0.15 per HIT and averaged 99 HITS per hour. This is more than prevailing local minimum wage. In total we paid \$5,400 to the crowd-sourced workers. They were told not to attempt the task if they did not wish to be exposed to offensive material.

The instructions and template for human evaluation are provided in Figure\ref{fig:human_ins} and Figure~\ref{fig:human_eval} below.
\begin{figure}[h!]
    \centering
    \includegraphics[scale=0.75]{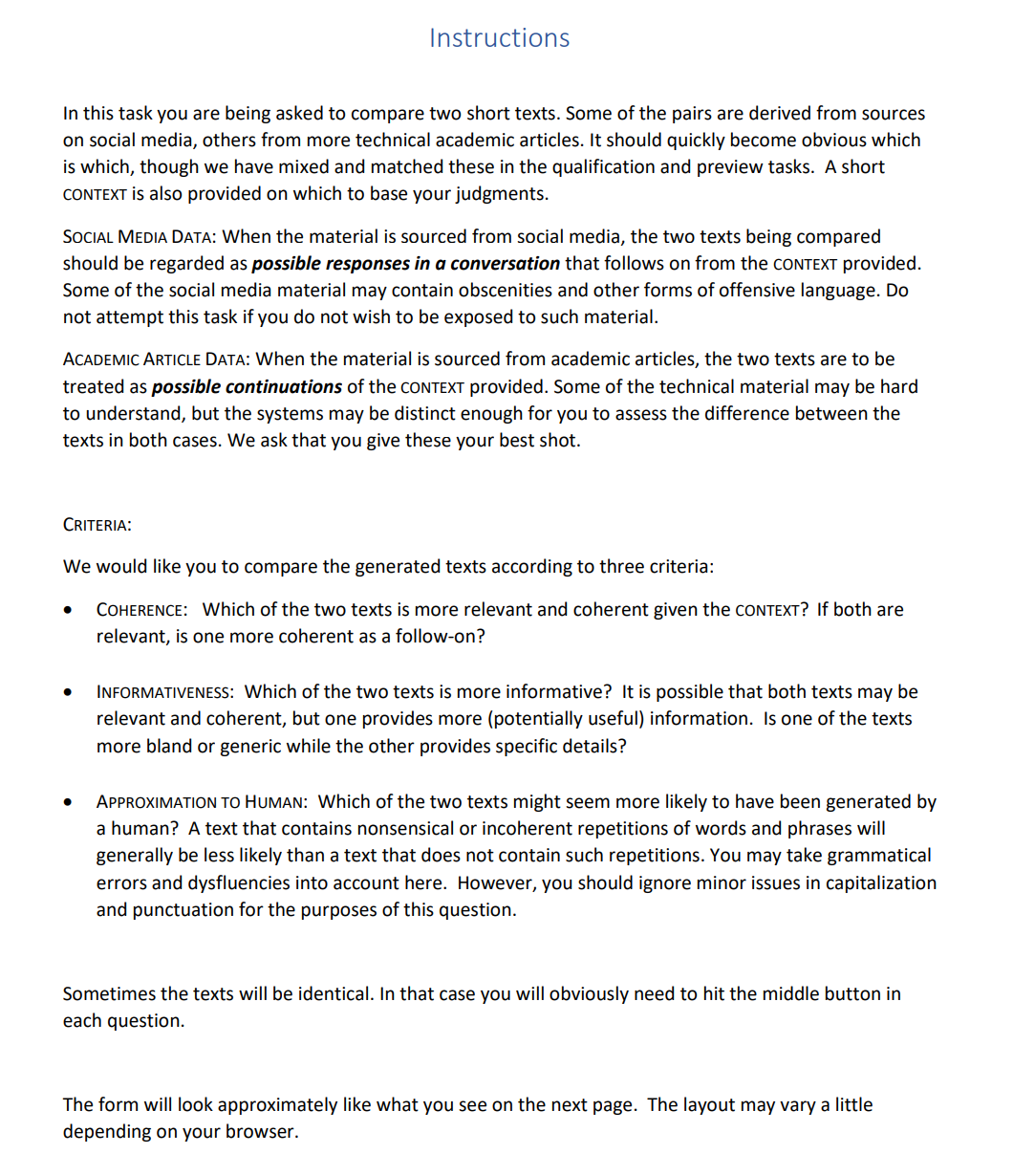}
    \caption{Human evaluation instructions.}
    \label{fig:human_ins}
\end{figure}

\begin{figure}[h!]
    \centering
    \includegraphics[scale=0.85]{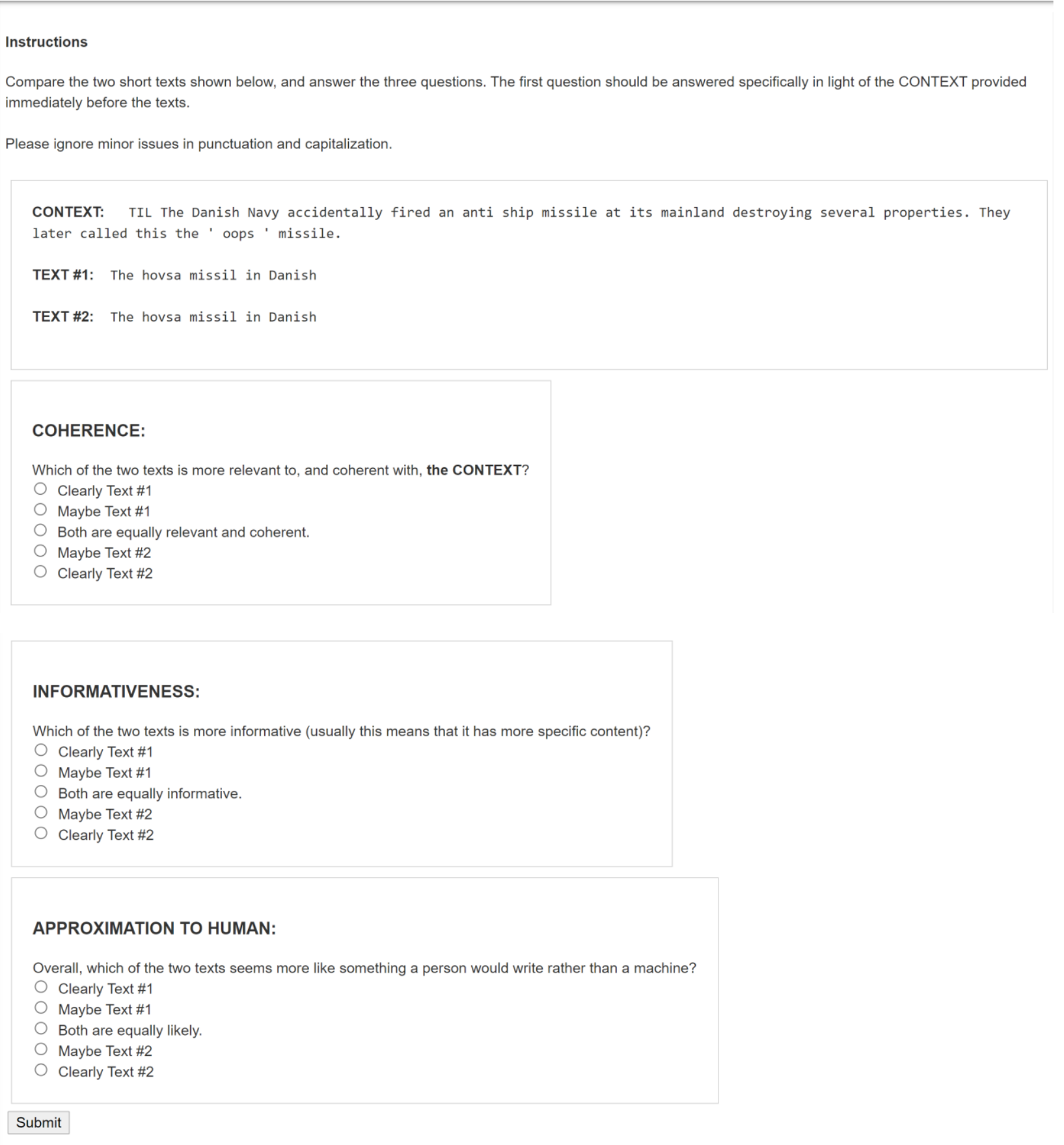}
    \caption{Human evaluation template.}
    \label{fig:human_eval}
\end{figure}

\end{document}